\pdfoutput=1
\documentclass[runningheads]{llncs}
\usepackage{graphicx}
\usepackage{bbding}
\usepackage{tikz}
\usepackage{comment}
\usepackage{amsmath,amssymb} 
\usepackage{color}

\definecolor{citecolor}{RGB}{34,139,34}
\usepackage{hyperref}
\hypersetup{breaklinks=true,letterpaper=true,colorlinks,citecolor=citecolor}
\PassOptionsToPackage{hyphens}{url}
\usepackage{hyperref}
\hypersetup{colorlinks=true}

\usepackage{multirow}
\usepackage{caption}
\newlength\savewidth
\usepackage{colortbl}
\usepackage{xcolor}
\definecolor{mygray}{gray}{.9}
\usepackage{booktabs}
\usepackage{xspace}

\newcommand{\app}{\raise.17ex\hbox{$\scriptstyle\sim$}}
\makeatletter\renewcommand\paragraph{\@startsection{paragraph}{4}{\z@}
  {.495em \@plus1ex \@minus.2ex}{-.5em}{\normalfont\normalsize\bfseries}}\makeatother

\makeatletter
\DeclareRobustCommand\onedot{\futurelet\@let@token\@onedot}
\def\@onedot{\ifx\@let@token.\else.\null\fi\xspace}

\def\eg{\emph{e.g}\onedot} 
\def\ie{\emph{i.e}\onedot} 
 
 \def\vs{\emph{vs}\onedot}

\makeatother

\begin{document}
\pagestyle{headings}
\mainmatter
\def\ECCVSubNumber{7899}  

\title{In Defense of  Image Pre-Training  for Spatiotemporal Recognition} 

\titlerunning{Image Pre-Training for Spatiotemporal Recognition}
%
\author{
Xianhang Li\textsuperscript{1} \and
Huiyu Wang\textsuperscript{2} \and
Chen Wei\textsuperscript{2} \and
Jieru Mei\textsuperscript{2} \and
Alan Yuille\textsuperscript{2} \and \\
Yuyin Zhou\textsuperscript{1} \and
Cihang Xie\textsuperscript{1}} 
\institute {\textsuperscript{1}University of California, Santa Cruz \quad \textsuperscript{2}Johns Hopkins University}

\authorrunning{X. Li et al.}
%

\maketitle
\begin{abstract}
Image pre-training, the current de-facto paradigm for a wide range of visual tasks, is generally less favored in the field of video recognition. 
By contrast, a common strategy is to directly train with spatiotemporal convolutional neural networks (CNNs) from scratch. Nonetheless, interestingly, by taking a closer look at these from-scratch learned CNNs,  we note there exist certain 3D kernels that exhibit much stronger appearance modeling ability than others, arguably suggesting appearance information is already well disentangled in learning.
Inspired by this observation, we hypothesize that the key to effectively leveraging image pre-training lies in the decomposition of learning spatial and temporal features, and revisiting image pre-training as the appearance prior to initializing 3D kernels.
In addition, we propose Spatial-Temporal Separable (STS) convolution, which explicitly splits the feature channels into spatial and temporal groups, to further enable a more thorough decomposition of spatiotemporal features for fine-tuning 3D CNNs. 

Our experiments show that simply replacing 3D convolution with STS notably improves a wide range of 3D CNNs without increasing parameters and computation on both Kinetics-400 and  Something-Something V2.
Moreover, this new training pipeline consistently achieves better results on video recognition with significant speedup. For instance, we achieve $+0.6\%$ top-1 of Slowfast on Kinetics-400 over the strong 256-epoch 128-GPU baseline while fine-tuning for only 50 epochs with 4 GPUs. The code and models are available at \url{github.com/UCSC-VLAA/Image-Pretraining-for-Video}.

\end{abstract}

\section{Introduction}
Deep convolutional neural networks (CNNs) pre-trained on large-scale datasets (\eg, ImageNet~\cite{imagenet2009imagenet}) play a vital role in computer vision.  
The spatial feature representations acquired by such models can then be transferred to the downstream task of interest via fine-tuning, leading to significant performance improvements especially for small target datasets.
This ``\emph{pre-training then fine-tuning}'' paradigm has gradually become the de-facto standard and established state-of-the-arts for a wide range of vision tasks, such as semantic segmentation \cite{fullycn2015,maskrcnn2017mask} and object detection \cite{fastrcnn2015,fatserrcnn2015}.

The ``\emph{pre-training then fine-tuning}'' paradigm is readily used for image recognition, as the two stages usually share the same backbone. However, for video recognition, to model the temporal dynamics and build spatiotemporal features, recent research efforts are dedicated to building video-specific architectures (\eg, Slowfast \cite{slowfast2019}, X3D \cite{x3d2020}).
However,
these structural changes prevent the model from getting image pre-training weights for \textit{free}.
One popular method to leverage image pre-training is to inflate 2D kernels pre-trained on ImageNet to 3D ones~\cite{k4002017}. 
However, recent studies such as SlowFast and Multigrid~\cite{slowfast2019,multigrid2020} suggest that using ImageNet pre-training does not outperform (sometimes even produces worse results than) training from scratch.
Hence,
the field of video representation learning has witnessed  a  paradigm  shift  from pre-training  then  fine-tuning to training  from scratch for today’s advanced spatiotemporal 3D CNNs.

As directly training these 3D CNNs from scratch requires enormous computational resources \cite{r2plus1d2018,slowfast2019,csn2019,multigrid2020,correlation2020,x3d2020},
many recent research efforts are devoted to designing efficient operations \cite{tsm2019,tea2020,nonlocal2018} and architectures \cite{r2plus1d2018,correlation2020} for 3D video recognition.
For instance, R2plus1D \cite{r2plus1d2018} is proposed to separate a 3D convolution into a 1D temporal convolution and a 2D spatial convolution.
Tran et al. propose CSN \cite{csn2019} which 
operates 3D convolution in the depth-wise manner.
This structural decomposition assumes that we should not treat space and time symmetrically.
However,
due to the end-to-end characteristic of training from scratch,
the spatiotemporal features are still jointly learned in these 3D CNNs in an unconstrained manner.
Thus, it remains unclear how spatial and temporal information is exploited in the network.

\begin{figure}[t!]
\centering
  \includegraphics[width=0.75\textwidth]{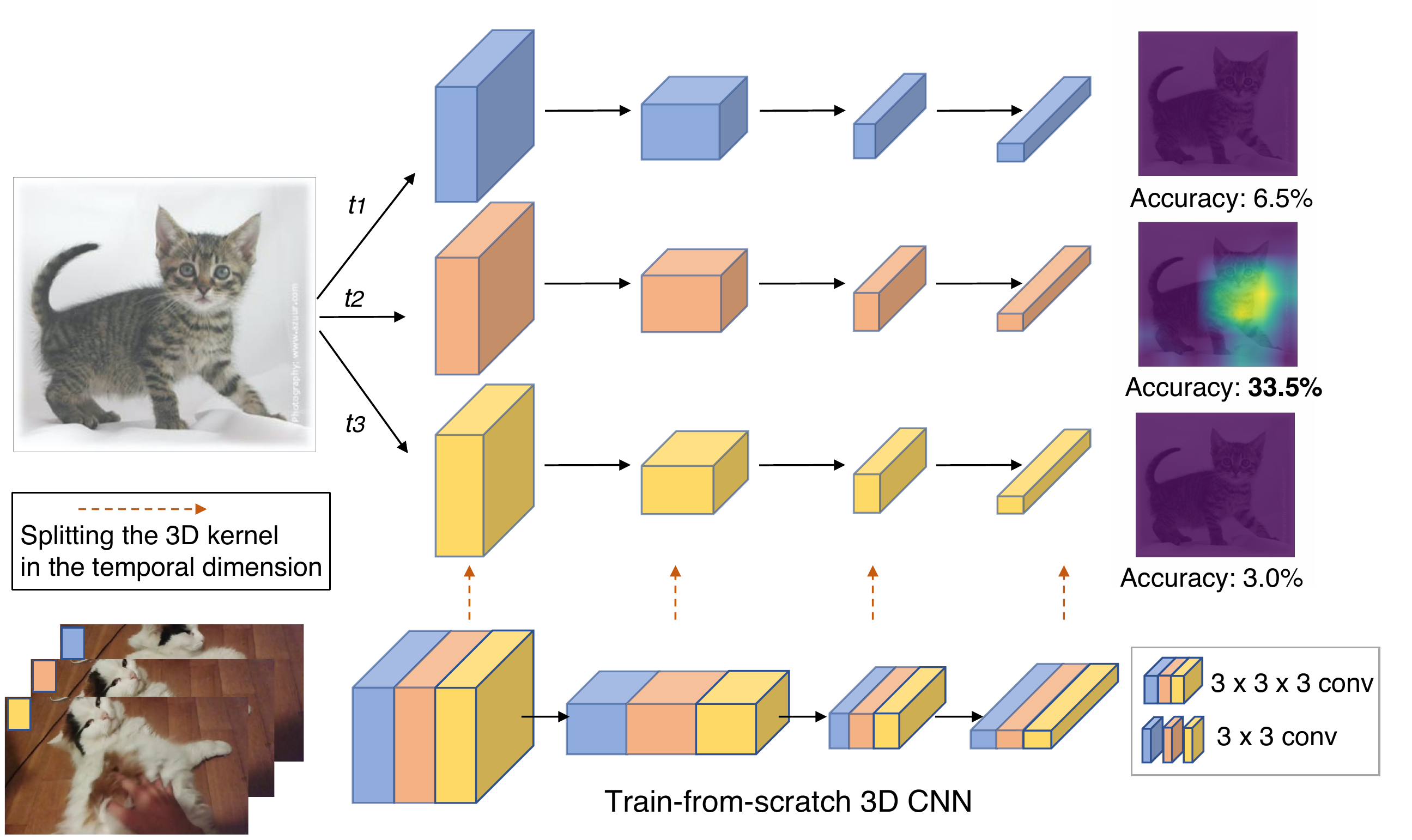}
\vspace{-0.5em}
\captionof{figure}{\textbf{Motivation}. By splitting the 3D convolution into several 2D ones along the temporal dimension, we can easily transform a 3D CNN into 2D CNNs. Note the resting 2D parts in the 3D CNN remain the same. Interestingly,  we note the 2D CNN along the center temporal direction attains much stronger linear-probing ImageNet accuracy than others, \ie, 33.5\% \vs 6.5\% or 3.0\%.}
\label{fig:motivation} 
\vspace{-1.8em}
\end{figure}

To better understand what these from-scratch trained 3D kernels learn, as shown in Figure~\ref{fig:motivation}, we split the 3D convolution trained on Kinetics-400 dataset~\cite{k4002017} using the CSN-50 network ~\cite{csn2019} into several 2D counterparts along the temporal dimension, and quantify the appearance
modeling power by assessing the linear probing performance of these 2D CNNs on ImageNet-1K.
Surprisingly, despite the domain gap between Kinetics-400 and ImageNet-1K, the 2D CNN along the center temporal direction (indicated by the orange branch in Figure~\ref{fig:motivation}) still achieves a non-trivial linear probing performance, indicating its strong appearance modeling ability. Moreover, 
using the gradients returned from logits in a category \cite{gradcam2017},  we plot the corresponding heat map that highlights the important regions in the image used for prediction, which are also highly correlated with the appearance information  (yellow and blue regions).

This interesting phenomenon further inspires us to rethink the value of image pre-training for 3D spatiotemporal recognition---if 3D kernels have such strong appearance modeling power, then the spatial features pre-trained from 2D images should be beneficial for training 3D CNNs.
As the appearance
information is already well disentangled in 3D CNNs, we hypothesize that the key to harnessing image pre-training lies in properly decomposing the spatiotemporal features learned from 3D kernels into spatial and temporal parts. 
Based on this principle, we design a new pre-training and fine-tuning pipeline to facilitate feature decomposition. 
Specifically, image pre-training is leveraged as the appearance prior for only initiating the 2D counterpart along the center temporal direction of the 3D convolution where the strongest appearance modeling power lies.
Then in the fine-tuning stage, we propose 3D Spatial-Temporal Separable (STS) convolution which explicitly splits the feature channels into spatial and temporal groups.
The former group learns the static appearance while the latter focuses on learning dynamic motion features, thus enabling a more thorough decomposition of spatiotemporal features for fine-tuning 3D CNNs.

Compared with prior solutions which train 3D CNNs from scratch, our method not only enjoys a significant speedup, but also improves 3D video representation learning by carefully tackling the appearance prior from image pre-training models.
We evaluate our method on a wide range of advanced 3D CNNs, including SlowFast~\cite{slowfast2019}, CSN~\cite{csn2019}, R2plus1D~\cite{r2plus1d2018}, and X3D~\cite{x3d2020}.
Without using more computation or parameters,
our method brings notable improvement on two popular video benchmarks Kinetics-400 \cite{k4002017} (\textbf{+2.1\%})  and Something-Something V2 \cite{sthv22017} (+\textbf{5.3\%}).
On Kinetics-400, our proposed training pipeline is at least \textbf{2$\times$} faster than training from scratch. 
The empirical analysis and design principles provided in our work will help researchers rethink the value of image pre-training and better understand its role in video recognition.

\section{Related Works}
\paragraph{Pre-training and fine-tuning for video classification.}
Pre-training on a large-scale dataset and then fine-tuning on a minor dataset for downstream tasks has been widely used to resist overfitting and get state-of-the-art performance \cite{twostream2014two,k4002017,fastrcnn2015,fatserrcnn2015,fullycn2015,maskrcnn2017mask}. This paradigm has also been employed and continuously discussed in video classification \cite{k4002017,spatiotemporal2016}. Due to the nature of temporal variance in the video, two types of methods can be used to accelerate the learning of video classification algorithms: image-based pre-training and video-based pre-training.  

\paragraph{Image pre-training.} Video classification models tend to consider both temporal and spatial information, and image pre-training can contribute to preserving spatial properties. There is active research about designing a well-performed video architecture based on a 2D ImageNet pre-trained backbone \cite{resnet2016d,vgg2014,inception2015g}. TSN \cite{tsn2016} is proposed to sample video frames sparsely that boost 2D image model performance on video classification. Revising a 2D block \cite{tei2020,tea2020,tsm2019,stm2019,ctnet2021,gateshift2020,smallbignet2020} or inserting attention-style block \cite{nonlocal2018,regiongraph2018,trn2018,globalreasoning2019,gcnet2019,disentangled2020} into a 2D model can enhance the ability of spatiotemporal representation. In order to make the 3D model converge faster, I3D \cite{k4002017} is proposed to initialize 3D convolution by inflating the 2D kernels. 
The recent success of the vision transformer \cite{vit2020,deit2021} has also made significant progress in video classification \cite{timesformer2021,vivit2021,videoswinvideo,mvt2021,li2022uniformer}, and using larger image datasets such as ImageNet-21K \cite{imagenet2009imagenet}, JFT-300M \cite{jft2017} significantly improves the model performance. Compared to these approaches, we find that the improvement from such pre-training strategies are still not well explored in 3D CNNs. 
Secondly, instead of using larger datasets, we focus on an more efficient and fair setting.


\paragraph{Efficient 3D Video Recognition.} 
Expanding a 2D image model is widely adopted in video recognition. First, extending the kernel into 3D increases the spatial and temporal receptive field and brings more parameters and computation \cite{c3d2015,p3d2017,s3d2018,movinet2021_CVPR}. Second, extending the temporal dimension into different resolutions further improves the performance. Recent work X3D \cite{x3d2020} considers all these expansions in a video-oriented perspective. 
Similarly, the decomposition can perform either from the model or the data perspective. R2plus1D \cite{r2plus1d2018} is proposed to separate the spatiotemporal convolution. There are other approaches which also separate the 3D convolution along the channel dimension for efficiency \cite{csn2019,ctnet2021}. From the data perspective, a two-stream architecture \cite{twostream2014two,spatiotemporal2016,slowfast2019,tpn2020} is a representative method that divides the input video into appearance and motion, \eg, optical flow \cite{twostream2014two} and temporal difference \cite{stm2019,tdn2021}. 
In this work, we focus on separating the learning process into efficient image pre-training and fine-tuning.
Furthermore, we proposed that the Spatiaotemporal Separable 3D convolution further improves the performance.

\section{Methodology}
In this section, we first reveal that in spatiotemporal recognition, 3D kernels still exhibit strong appearance modeling ability (Section~\ref{sec:motivation}).
This observation further inspires us to design a novel pre-training \& fine-tuning paradigm for video recognition, where 2D image pre-training serves the appearance prior, guiding the further spatiotemporal fine-tuning process (Section~\ref{sec:pretrain_and_finetune}).
Finally in Section \ref{sec:budget}, we provide two general training settings to conduct a fair comparison with training from scratch.

\subsection{Appearance Modeling in 3D Kernels}
\label{sec:motivation}
\begin{table}[t!]
    \centering
    \resizebox{0.5\linewidth}{!}{
    \begin{tabular}{c|c|c|c}
    \toprule
    &t1-CNN  &t2-CNN  &t3-CNN      \\
    \midrule
    \midrule
    CSN-50 \cite{csn2019}  &$3.0$  &$\textbf{33.5}$        &$6.3$    \\ 
    X3D-S \cite{x3d2020}  &$5.7$    &$\textbf{41.2}$     &$5.5$  \\ 
    R2plus1D-34 \cite{r2plus1d2018} &$0.8$   &$\textbf{32.3}$     &$2.2$   \\  
    Slowonly-50 \cite{slowfast2019}  &$18.3$  &$\textbf{45.1}$   &$17.6$ \\ 
    \bottomrule
    \end{tabular}
    }
    \vspace{0.5em}
    \captionof{table}{ImageNet-1K \textbf{Linear Probing} Top1 Accuracy. The weights come from the models trained from scratch on Kinetics-400. We fine tune the linear classification layer for 100 epochs on ImageNet-1K.}
    \label{tab:motivation}
    \vspace{-3em}
\end{table}

Despite numerous variants of 3D CNNs being proposed, it is unclear what these 3D kernels actually learn. 
In this paper, to better understand the properties of 3D kernels, we aim to examine the relationship between the 3D kernels and 2D appearance modeling by answering the following question: \emph{Do 3D features equally model the 2D appearance across the temporal dimension?}

To answer this question, we propose to split the standard 3D convolution into several 2D ones along the temporal dimension, and quantify the appearance modeling of the 3D CNN by assessing the linear probing performance of these 2D CNNs instead.
Specifically, we split a 3D convolution ($3 \times3 \times 3$ or $3 \times1 \times 1$) into three 2D convolutions ($3 \times3 $ or $1 \times1$) at different time stamps t1, t2, t3, as shown in Figure \ref{fig:motivation}. 
We then evaluate these 2D CNNs on ImageNet-1k by freezing the weights and only fine-tuning the linear classification layer. 
As shown in Table~\ref{tab:motivation}, we examine four widely-used video classification models: 1) CSN-50 \cite{csn2019}, 2) X3D-S \cite{x3d2020}, 3) R2plus1D-34 \cite{r2plus1d2018},
4) Slowonly-50 \cite{slowfast2019}, all of which are trained from scratch on Kinetics-400.  
Surprisingly, though there is a domain gap between Kinetics-400 and ImageNet-1k, out of the three 2D CNNs, the middle one at t2 consistently achieves a non-trivial linear probing performance, whereas in the other two 2D networks at t1 and t3 time stamps, we observe severe performance degradation. 
This demonstrates that while spatiotemporal features are not equivalent in terms of appearance modeling, they do exhibit strong appearance modeling ability.

This phenomenon further motivates us to rethink the value of image pre-training in spatiotemporal recognition.
Specifically, \emph{can we achieve a better trade-off between performance and training efficiency}?
Next, we will elaborate on how image pre-training can be leveraged to improve spatiotemporal recognition.

\begin{figure}[t!]
\centering
\includegraphics[width=0.95\linewidth]{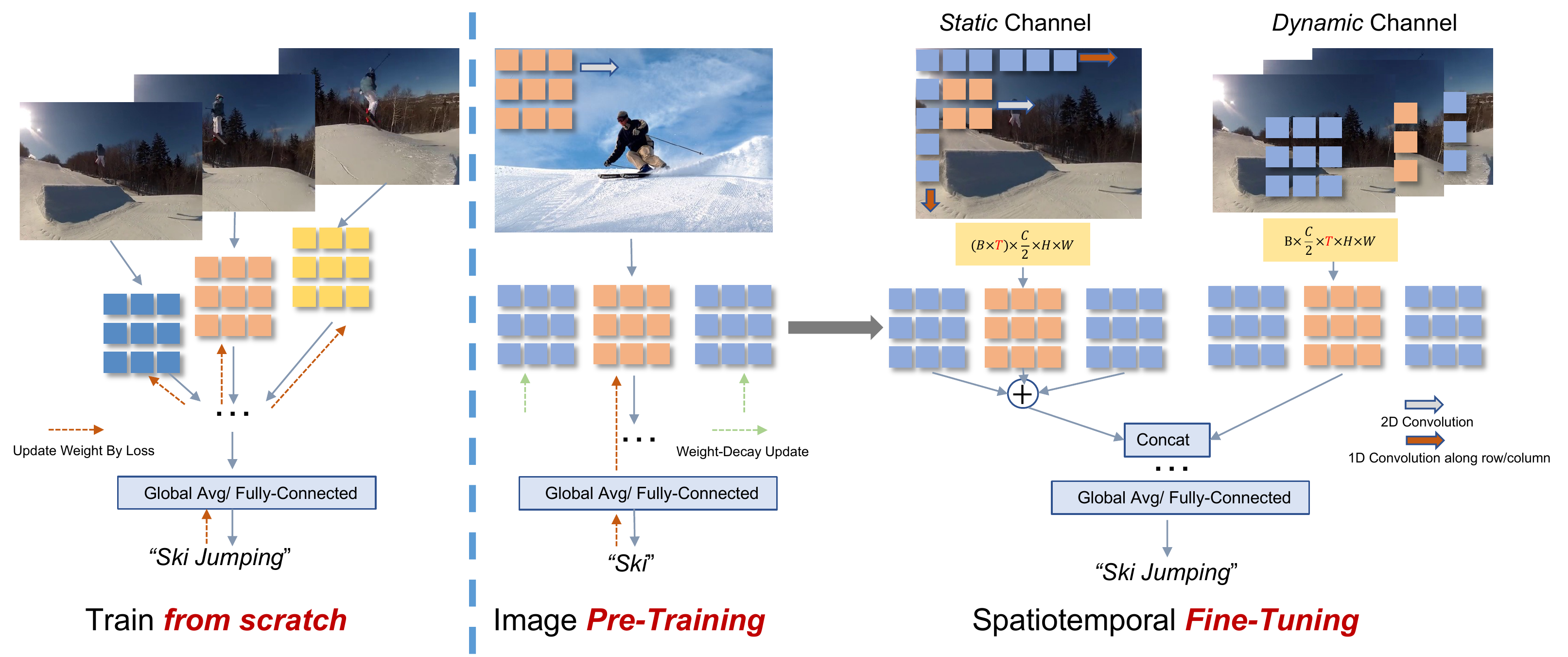}
\vspace{-0.2em}
\caption{\textbf{Overview of Image Pre-Training \& Spatiotemporal Fine-Tuning.} Left: training from scratch learns features in a joint manner. Right: decompose the learning into image pre-training and spatiotemporal fine-tuning. }
\label{fig:method} 
\vspace{-2em}
\end{figure}

\subsection{Image Pre-Training and Spatiotemporal Fine-tuning}
\label{sec:pretrain_and_finetune}

The value of ImageNet pre-training is firstly challenged in  He et al. ~\cite{rethinking2019}.
Later, many recent studies have further demonstrated that compared with training from scratch, ImageNet pre-training even results in performance drop for video action recognition models~\cite{x3d2020,csn2019,r2plus1d2018,slowfast2019}. 
However, as shown in Section~\ref{sec:motivation}, by closely examining these from-scratch learned 3D kernels, we find they still exhibit strong appearance modeling power.
Based on this interesting observation, we hypothesize that image pre-training is indeed valuable for training video recognition models, when properly decomposing the learning of spatiotemporal features.
To this end, we propose a novel pre-training and fine-tuning paradigm.
As shown in Figure \ref{fig:method}, compared with the popular train-from-scratch pipeline which learns spatial and temporal features in a joint fashion, our proposed method thoroughly decomposes the spatial and temporal features via the following two steps: 1) image pre-training for appearance modeling, and 2) spatiotemporal fine-tuning.

\paragraph{Image pre-training for appearance modeling.} 
Based on our observation in Section~\ref{sec:motivation} that 3D kernels usually exhibit strong appearance ability along the center temporal direction, we propose to leverage image pre-training as the appearance prior by only initiating the middle kernel of the 3D convolution. As other parts of the 3D convolution do not have a strong relationship with appearance modeling, the rest of the weights will simply be updated by weight-decay regularization, as shown in Figure \ref{fig:method}. Leveraging the appearance prior from image pre-training plays a key role in facilitating the following fine-tuning stage since 1) it provides better initialization for the 3D kernels, and 2) it helps significantly reduce the fine-tuning schedule.
Given that the appearance modeling is already well established, the 3D CNN only needs to focus on learning the temporal information.

\paragraph{Spatiotemporal fine-tuning.} 
After acquiring the appearance prior from image pre-training, we then present a novel spatiotemporal fine-tuning method for a more thorough decomposition of spatiotemporal features. 
Specifically, as shown in Figure \ref{fig:method}, we propose 3D Spatiotemporal Separable (STS) convolution, which splits the 3D convolution along the output channel dimension into two groups, one for the static appearance modeling, and the other for dynamic motion modeling. The motion group builds the spatiotemporal feature by a 3D convolution, while the appearance group focuses on the spatial feature.
In the appearance group, we divide the original 3D convolution along the temporal dimension into one 2D convolution (\eg, 3$\times$3) and two 1D convolutions (\eg, 1$\times$9 and 9$\times$1), to enlarge the receptive field.  The formal definition of STS can be found  in the Appendix.
The reason is that if we directly apply a 3D convolution on the spatial feature (\eg, \textbf{3}$\times$3$\times$3), the receptive field is limited to the middle 2D spatial convolution (\textbf{1}$\times$3$\times$3).
By contrast, in the proposed STS, 
the 1D convolutions can build the spatial feature along each row/column as shown in Figure \ref{fig:method}, which enlarges
the original receptive field in the 2D space (3$\times$3) to be (3$\times$3+1$\times$9+9$\times$1) without increasing parameters and computation.
With a strong appearance prior,
STS can significantly boost the performance during the fine-tuning stage benefiting from the more discriminative spatial features.

\begin{table}[t!]
    \centering
    \resizebox{0.95\linewidth}{!}{
    \begin{tabular}{c|c|c|c|c}
    \toprule
    & \# Instance  & \# Images (frames)  &Training Epochs
     &Total Training Images\\
    \midrule
    \midrule
    ImageNet-1K             &$1.28M$  &$1.28M$        &X   &$X \times 1.28M$\\ 
    Kinetics-400  &$0.24M$    &$68.89M$    &Y  &$Y \times T \times 0.24M$\\ 
    Something-Something V2 &$0.17M$   &$18.98M$     &Z  &$Z \times T \times 0.17M$ \\  
    \bottomrule
    \end{tabular}
    }
    \vspace{0.5em}
    \captionof{table}{\textbf{Dataset Settings.} Each video clip is an instance. Total training images are total number of instances multiply the number of input frames T.}
    \label{tab:dataset images}
    \vspace{-1.3em}
\end{table}

\begin{figure}[t!]
\vspace{-1em}
\centering
\includegraphics[width=\linewidth]{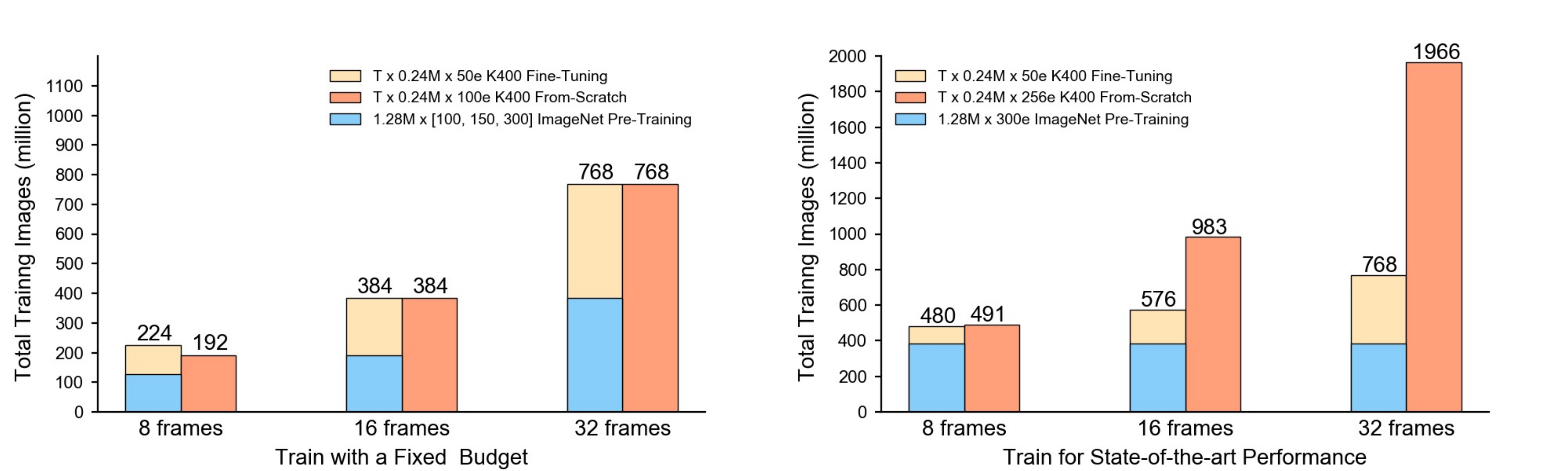}
\vspace{-1.8em}
\caption{\textbf{Computation Budget Towards Different Input Frames.}  Total numbers of images/frames seen during all training epochs, for pre-training + fine-tuning (blue + yellow
bars) \vs from scratch (orange bars). In Fixed-Budget setting (left), we pre-train for 100, 150 and 300 epochs to match total computation. In the train-for-state-of-the-art setting (right), the pre-train is 300 epoch. In both setting, we fine-tune for 50 epochs.}
\label{fig:computation budget} 
\vspace{-1.3em}
\end{figure}

\subsection{Training with a Fair Budget}
\label{sec:budget}
Following the fair comparison setting \cite{rethinking2019} between pre-training \& fine-tuning and training from scratch, we consider total training images as the condition for controlled experiments.
We show the comparison on different datasets in Table \ref{tab:dataset images}.
A typical training from scratch schedule on Kinetics-400 in recent advanced spatiotemporal 3D CNNs \cite{slowfast2019,x3d2020} involves $T\times0.24M$ frames iterated for 256 epochs.  
For example, Slowfast\cite{slowfast2019} that take 32 frames as input for each training epoch will ``see" over \textbf{7.68} million images, which is \textbf{$\times$6} more than one epoch ImageNet training. 
Compared with the training from scratch schedule,
the pre-training \& fine-tuning schedule (90 epochs pre-training + 100 epochs fine-tuning) consumes much less computation, which is usually unfair.
As shown in Figure \ref{fig:computation budget},
we provide two general settings about how to fairly compare the performance of pre-training \& fine-tuning and training from scratch.

\paragraph{Training with a fixed budget.}
Compared with training from scratch, we aim to maximize the performance of pre-training and fine-tuning with the same amount of computation. In this setting, we allocate around half of the total training budget to image pre-training, and another half to spatial-temporal fine-tuning, as shown in Figure \ref{fig:computation budget}. Surprisingly, we found that this simple strategy works well no matter the model consumes 8 frames, 16 frames, or 32 frames, according to our experiments.

\paragraph{Training for state-of-the-art.} 
In this setting, we provide a more efficient training pipeline to match the previous state-of-the-art CNNs performance. Instead of scaling the image pre-training budget with respect to the total budget, we train our model with images for a long schedule (\eg, 300 epochs) until convergence. As shown in Figure~\ref{fig:computation budget}, we then fine-tune our model with a fixed short schedule (50 epochs) for all frame settings. We notice that the total computation cost is still lower than training on videos from scratch but the performance can be significantly improved over a wide range of advanced 3D CNNs on Kinetics-400 (\textbf{+2.1\%}) and SS-V2 (\textbf{+4.1\%}).

\section{Experiments and Results}

\paragraph{Datasets.} We choose ImageNet-1K \cite{imagenet2009imagenet} as our primary image pre-training dataset. 
It contains around 1.28 million images for training and $50K$ images for validation.
We also compare different pre-training datasets in the ablation study, \eg, Kinetics-400.  
Unless otherwise specified, our default pre-training dataset is ImageNet-1K. 
Then we fine-tune our models on widely-used Kinetics-400 (abbreviated as K400) \cite{k4002017} and Something-Something V2 (abbreviated as SS-V2) \cite{sthv22017}. 
Kinetics-400 contains about $260K$ videos of $400$ different human action categories. We use the training split ($240K$ videos) for training and the validation split ($20K$ videos) for evaluating different models. SS-V2 contains $220K$ videos of 174 predefined human-object interactions with everyday objects.

\paragraph{Training.} We pre-train all models for 100, 150, 300 epochs, respectively, based on different computation budgets. 
In the 100 and 150 epochs schedule, the model is trained using SGD with a learning rate of 0.4, a momentum of 0.9, a weight decay of 0.0001, and a batch size of 1024. The learning schedule is cosine annealing with a 5-epoch warm-up, and the initial learning rate is 1e-6. For augmentation, we enable the RandAugment \cite{randaugment2020}, Mixup \cite{mixup2017}, and Cutmix \cite{cutmix2019}. We only use label smoothing as regularization. 
In the 300 epochs setting, we follow the training recipe in \cite{resnetstrikeback2021}. In the next fine-tuning stage, we use a dense-sampling strategy for Kinetics-400 \cite{slowfast2019,x3d2020} and an n-segments based method sampling strategy for SS-V2 \cite{tsn2016,tsm2019}. 
During training, random resize cropping is utilized for data augmentation, and the cropped region is resized to 224 × 224 for each frame. 
For K400,
the batch size, initial learning rate, weight decay, and dropout rate are set as 128 or 64 (based on input frames),  0.02 or 0.01, 1e-4, and 0.5 respectively.
For SS-V2,
we change the learning rate to 0.1 and the batch size is 64. 
Unless otherwise specified, 
the networks are fine-tuned for 50 epochs using SGD.

\paragraph{Inference.} For K400, following the common practice \cite{nonlocal2018,slowfast2019,x3d2020} we uniformly sample 10 clips from a video along its temporal axis. We scale the shorter spatial axis to 256 pixels for each clip and take 3 crops of 256$\times$256 to cover the spatial dimensions. We average the softmax scores for prediction.
For SS-V2, we only sample 1 clip based on the n-segments method and conduct the 3-crop testing following the common recipe in \cite{tsm2019,ctnet2021}.

\subsection{Comparison on Common 3D CNN Architectures}

\begin{table}[t!]
    \centering
    \resizebox{0.98\linewidth}{!}{
    \begin{tabular}{lc|c|c|c|c|cc}
    \toprule
      \multirow{2}{*}{}&+ Image Pre. & Pre-Train  &  Fine-tune &\multirow{2}{*}{Total Time} &\multirow{2}{*}{Speedup} &\multirow{2}{*}{K400} &\multirow{2}{*}{$\mathbf{\Delta}$}  \\
      & \& STS Conv & (hrs)  & (hrs)  &  &  & & \\
    \midrule
    \midrule
    \multirow{2}{*}{Slowonly50-8\textit{f} \cite{slowfast2019}}   
     &\XSolidBrush &-  &389 &389  & $\times1$  &74.9 & -  \\ 
     & \cellcolor{mygray} \CheckmarkBold   & \cellcolor{mygray} 182   & \cellcolor{mygray}\textbf{76}       &$ \cellcolor{mygray} \textbf{258}$  & \cellcolor{mygray} $\times \mathbf{1.5}$   & \cellcolor{mygray}\textbf{75.6}  & \cellcolor{mygray}\textbf{+0.7}\\
    \midrule
    \multirow{2}{*}{Slowfast50 (4x16) \cite{slowfast2019}} 
     &\XSolidBrush &-   &696       &$  696$ & $\times1$  &75.6  &-\\ 
     &\cellcolor{mygray} \CheckmarkBold  & \cellcolor{mygray} 237   & \cellcolor{mygray} \textbf{136}      & \cellcolor{mygray} $\textbf{373}$ & \cellcolor{mygray} $\times \mathbf{1.9}$  & \cellcolor{mygray} \textbf{76.2}  & \cellcolor{mygray} \textbf{+0.6}\\
    \midrule
    \multirow{2}{*}{Slowfast50 (8x8) \cite{slowfast2019}} 
     &\XSolidBrush &-   &840       &$ 840$ & $\times1$  &76.9  &-\\ 
     &\cellcolor{mygray} \CheckmarkBold  & \cellcolor{mygray} 237   & \cellcolor{mygray} \textbf{164}      &\cellcolor{mygray} $\textbf{401}$ & \cellcolor{mygray} $\times \mathbf{2.1}$  & \cellcolor{mygray} \textbf{77.2}  & \cellcolor{mygray} \textbf{+0.3}\\
    \midrule
    \multirow{2}{*}{R2plus1D34-8\textit{f} \cite{r2plus1d2018}}
     &\XSolidBrush  &-    &778  &$778$ & $\times1$  &68.7 &-\\
     &\cellcolor{mygray} \CheckmarkBold   &\cellcolor{mygray} 230    & \cellcolor{mygray} \textbf{152}  & \cellcolor{mygray} $\textbf{382}$ & \cellcolor{mygray} $\times \mathbf{2.0}$    &\cellcolor{mygray} \textbf{76.2} & \cellcolor{mygray} \textbf{+7.5}\\
    \midrule
    \multirow{2}{*}{CSN50-32\textit{f} \cite{csn2019}}
     &\XSolidBrush &-        &1106   &$1106$  & $\times1$  &73.6 &-\\ 
     &\cellcolor{mygray} \CheckmarkBold   &\cellcolor{mygray} 176        &\cellcolor{mygray} \textbf{216}  &\cellcolor{mygray} $ \textbf{392}$ & \cellcolor{mygray} $\times \mathbf{2.8}$  & \cellcolor{mygray} \textbf{76.7} & \cellcolor{mygray} \textbf{+3.1}\\
    \midrule
    \multirow{2}{*}{X3D-S-13\textit{f} \cite{x3d2020}} 
    
     &\XSolidBrush &-   &432     &432 & $\times1$  &73.2 &-\\ 
     &\cellcolor{mygray} \CheckmarkBold  & \cellcolor{mygray} 121   &\cellcolor{mygray} \textbf{144}     &\cellcolor{mygray} $\textbf{265}$ & \cellcolor{mygray} $\times \mathbf{1.6}$  & \cellcolor{mygray} \textbf{73.5} & \cellcolor{mygray}\textbf{+0.3}\\
    \bottomrule
    \end{tabular}
    }
    \vspace{0.5em}
    \captionof{table}{\textbf{Comparing Top-1 Accuracy of Common 3D CNNs on K400.} Training time is measured in GPU hours on the same hardware and software implementation for all methods. Baseline performance is obtained by testing the released model. We pre-train all models for 300 epochs on ImageNet.}
    \label{tab:main results k400}
    \vspace{-3em}
\end{table}

\paragraph{Main results on K400.} We first demonstrate the evaluation of common 3D CNNs on K400 in Table \ref{tab:main results k400}. 
Firstly, the image pre-training (denoted by \textbf{Image Pre.}) and the proposed STS (denoted by \textbf{STS Conv}) both improve the video classification accuracy on a wide range of advanced video models, including Slowonly \cite{slowfast2019}, Slowfast \cite{slowfast2019}, R2plus1D \cite{r2plus1d2018}, CSN \cite{csn2019} and X3D \cite{x3d2020}, 
by an average of $\textbf{2.1 \%}$. 
Note that all models are trained with the same pre-training and fine-tuning pipeline (300 epochs pre-training + 50 epochs fine-tuning) described in Section \ref{sec:budget}, except for X3D-S. 
Due to its extremely tiny characteristic we extend the fine-tuning epochs for X3D-S to 100. 
These results demonstrate the strong generalization ability of our proposed pre-training and fine-tuning pipeline.
Secondly,
regarding the training efficiency,
our method is ${\times \textbf{2}}$ faster than training from scratch baseline on average.
Note that to be as fair as possible, we take the pre-training time into account of the total training time.
In addition, all methods compared in Table \ref{tab:main results k400} use the same hardware and software implementation.
More importantly,
the training-from-scratch recipe usually includes a large batch size and long training schedule,
which requires significant computation resources.
For example,
X3D-S \cite{x3d2020}, Slowfast \cite{slowfast2019} are trained on 128-GPU and R2plus1D-34 \cite{r2plus1d2018}, CSN-50 \cite{csn2019} are trained on 64-GPU cluster.
In our experiments,
with our pre-training and fine-tuning pipeline,
we can fine-tune the model with a much smaller batch size and achieve better results.
Thus,
our method requires significantly fewer computational resources. 
All results are obtained on a single 4-GPU machine.

\begin{table}[t!]
    \centering
    \resizebox{0.90\linewidth}{!}{
    \begin{tabular}{l|cc|c|c|cc}
    \toprule
    & Parameters  &GFLOPs  &Image Pre. &STS Conv    &SS-V2 &$\mathbf{\Delta}$  \\
    \midrule
    \midrule
    \multirow{3}{*}{Slowonly50-8\textit{f} \cite{slowfast2019}} 
    &\multirow{3}{*}{32.0M}   & \multirow{3}{*}{54.9} &\XSolidBrush &\XSolidBrush   &58.4 & -  \\
    &&&\CheckmarkBold &\XSolidBrush  &61.6 & \textbf{+3.2}  \\ 
    &&&\CheckmarkBold &\CheckmarkBold   &\textbf{62.7}  &\textbf{+4.3}\\
    \midrule
    \multirow{3}{*}{Slowfast50 (4x16)-32\textit{f} \cite{slowfast2019}}
    &\multirow{3}{*}{34.0M}    & \multirow{3}{*}{36.4} &\XSolidBrush &\XSolidBrush    &49.9  &-\\
    &&&\CheckmarkBold &\XSolidBrush   &55.8  &\textbf{+5.9}\\ 
    &&&\CheckmarkBold &\CheckmarkBold    &\textbf{57.2}  &\textbf{+7.3}\\
    \midrule
    \multirow{3}{*}{R2plus1D34-8\textit{f} \cite{r2plus1d2018}} 
    &\multirow{3}{*}{63.6M}  & \multirow{3}{*}{114.8} &\XSolidBrush &\XSolidBrush   &59.1 &-\\
    &&&\CheckmarkBold &\XSolidBrush   &62.3 &\textbf{+3.2}\\
    &&&\CheckmarkBold &\CheckmarkBold   &\textbf{63.0} &\textbf{+3.9}\\
    \midrule
    \multirow{3}{*}{CSN50-8\textit{f} \cite{csn2019}} 
    &\multirow{3}{*}{12.7M}  &\multirow{3}{*}{26.6}  &\XSolidBrush &\XSolidBrush  &57.5 &-\\
    &&&\CheckmarkBold &\XSolidBrush  &60.4 &\textbf{+2.9}\\ 
    &&&\CheckmarkBold  &\CheckmarkBold   &\textbf{61.4} &\textbf{+4.0}\\
    \midrule
    \multirow{3}{*}{X3D-S-8\textit{f} \cite{x3d2020}}
    &\multirow{3}{*}{3.3M}   &\multirow{3}{*}{3.2} &\XSolidBrush &\XSolidBrush  &51.9  & -\\
    &&&\CheckmarkBold &\XSolidBrush &57.0  &\textbf{+5.1}\\ 
    &&&\CheckmarkBold &\CheckmarkBold   &\textbf{58.3} &\textbf{+6.4}\\
    \bottomrule
    \end{tabular}
    }
    \vspace{0.5em}
    \captionof{table}{\textbf{Comparison on Common 3D CNNs on SS-V2.} We show the comparison on training from scratch, image pre-training and applying STS convolution respectively. We use test resolution (T$\times$256$\times$256) to measure GFLOPS.}
    \label{tab:main results SSV2}
    \vspace{-3em}
\end{table}

\paragraph{Main results on SS-V2.} 
SS-V2 is relatively small, therefore training efficiency is not our goal on this dataset.
Here we only use the fixed budget setting as described in Section \ref{sec:budget} and fairly train all the methods in our machine within the same setting.
We evaluate the temporal modeling ability of our proposed training pipeline and the STS convolution respectively over five common 3D video models in Table \ref{tab:main results SSV2}.
First, 
SS-V2 is known as a temporal-related dataset. 
The appearance modeling is considered less important than temporal modeling \cite{tsm2019,stm2019,tea2020}.
It is interesting to see that the image pre-training and fine-tuning improve the training from scratch methods by a large margin of $\textbf{4.1\%}$.
The significant improvements demonstrate the superiority of our proposed training pipeline.
Second,
the STS convolution can replace any 3D convolution layer (\eg, 3$\times$3$\times$3 or 3$\times$1$\times$1 ) without increasing parameters and computation.
Upon image pre-training,
using STS convolution can boost the performance by an average of $\textbf{1.2\%}$,
which clearly shows that the STS convolution is more suitable for temporal modeling.
Equipped with both proposed methods, 
we in total improve the baseline methods by an average of $\textbf{5.3\%}$,
which indicates that the STS convolution benefits more from the proposed training pipeline.

\section{Ablation Study}

\newcommand{\blocka}[3]{\multirow{3}{*}{\(\left[\begin{array}{c}\text{1$\times$1$ \times$1, #1}\\[-.1em] \text{1$\times$3$ \times$3, #2}\\[-.1em]\text{1$\times$1$ \times$1, #1}\end{array}\right]\)$\times$#3}}
\newcommand{\blockc}[3]{\multirow{3}{*}{\(\left[\begin{array}{c}\text{3$\times$1$ \times$1, #1}\\[-.1em] \text{1$\times$3$ \times$3, #2}\\[-.1em]\text{1$\times$1$ \times$1, #1}\end{array}\right]\)$\times$#3}}

\newcommand{\blockb}[3]{\multirow{3}{*}{\(\left[\begin{array}{c}\text{1$\times$1$ \times$1, #1}\\[-.1em] \text{3$\times$3$ \times$3, #2}\\[-.1em]\text{1$\times$1$ \times$1, #1}\end{array}\right]\)$\times$#3}}

Next, we fairly perform an in-depth experimental analysis and complete ablation experiments on our proposed training pipeline and STS convolution. 
We use the fixed budget setting as default pipeline.
All the from-scratch baselines are trained for 100 epochs.
When evaluating our methods, we only fine-tune 50 epochs.

\begin{table}[t!]
\begin{center}
\resizebox{0.85\linewidth}{!}{
\begin{tabular}{c|c|c|c}
\toprule
layer name & output size & ResNet50-3$\times$1$\times$1 & ResNet50-3$\times$3$\times$3* \\
\midrule
\midrule
conv1 & T$\times$112$\times$112 & \multicolumn{2}{c}{1$\times$7$\times$7 or 3$\times$7$\times$7, 64, stride 1$\times$2$\times$2}\\
\midrule
pool1 & T$\times$56$\times$56 & \multicolumn{2}{c}{max, 1$\times$3$\times$3, stride 1$\times$2$\times$2}\\
\midrule
\multirow{3}{*}{stage 1} & \multirow{3}{*}{T$\times$56$\times$56} & \blocka{$256$}{$64$}{$3$} & \blockb{$256$}{$64$}{$3$}\\
  &  &  & \\
  &  &  & \\
\midrule
\multirow{3}{*}{stage 2} &  \multirow{3}{*}{$T\times$28$\times$28}  & \blocka{$512$}{$128$}{$4$}  & \blockb{$512$}{$128$}{$4$}  \\
  &  &  & \\
  &  &  & \\
\midrule
\multirow{3}{*}{stage 3} & \multirow{3}{*}{$T\times$14$\times$14}  & \blockc{$1024$}{$256$}{$6$} & \blockb{$1024$}{$256$}{$6$} \\
  &  &  & \\
  &  &  & \\
\midrule
\multirow{3}{*}{stage 4} & \multirow{3}{*}{$T\times$7$\times$7}  & \blockc{$2048$}{$512$}{$3$} & \blockb{$2048$}{$512$}{$3$} \\
  &  &  & \\
  &  &  & \\
\midrule
pool5 & 1$\times$1$\times$1  & \multicolumn{2}{c}{spatial-temporal avg pool, fc layer with softmax} \\
\bottomrule
\end{tabular}
}
\end{center}
\vspace{-0.5em}
\caption{\textbf{ResNet3D architectures considered in our exploration experiments}. The backbone is ResNet50 \cite{resnet2016d}. Note that we set the temporal downsampling rate is 1 by default. *: to reduce computation and parameters, all the 3$\times$3$\times$3 convolutions are channel-wise.}
\label{tab:basic architecture}
\vspace{-3em}
\end{table}

\noindent\textit{Base architecture.}  In general, 3D convolution can be divided into temporal 3D convolution and spatial-temporal 3D convolution. We chose two corresponding architectures to verify the generality and effectiveness of our method,  presented in Table \ref{tab:basic architecture}.   Their backbone is ResNet 50 \cite{resnet2016d}. They are distinguished into ResNet50-3$\times$1$\times$1 and ResNet50-3$\times$3$\times$3 depending on whether the convolution of the core bottleneck block is $3\times1\times1$ or $3\times3\times3$. It is worth noting that all the 3D convolutions in ResNet50-3$\times$3$\times$3 are channel-wise to reduce the computation and the number of parameters. Following Slowfast and X3D \cite{slowfast2019,x3d2020}, the temporal downsampling rate is always set to 1. Specifically, our model takes clips with a size of T$\times$224$\times$224 where $T=\{8, 16, 32\}$ is the number of frames.

\subsection{Ablation on Image Pre-Training }
\begin{figure}[h!]
\vspace{-0.5em}
\begin{minipage}{\linewidth}
  \begin{minipage}[t]{0.48\linewidth}
    \centering
    \resizebox{\linewidth}{!}{
    \begin{tabular}{c|c|c|c}
    \toprule
    ResNet50-\textbf{3x1x1}    &K400 
      & SS-V2  & Training Budget \\
    \midrule
    \midrule
    From Scratch   &73.1        &58.4   & $ 192M$ \\ 
    \midrule
    \rowcolor{mygray}
    Pre-train\&Fine-tune      &$\textbf{74.9}$   &$\textbf{61.6}$  &$\times 1.16$\\ 
    \rowcolor{mygray}
       + STS Conv    &$\textbf{75.0}$ &$\textbf{62.7}$  &$\times 1.16$ \\ 
    \bottomrule
    \end{tabular}
    }
    \vspace{-0.5em}
    \label{tab:effectiveness 3x1x1}
  \end{minipage}
  \hfill
  \begin{minipage}[t]{0.48\linewidth}
    \centering
    \centering
    \resizebox{\linewidth}{!}{
    \begin{tabular}{c|c|c|c}
    \toprule
    ResNet50-\textbf{3x3x3}     &K400 
      & SS-V2  & Training Budget \\
    \midrule
    \midrule
    From Scratch    &70.7        &$57.5$   & $ 192M$ \\ 
    \midrule
    \rowcolor{mygray}
    Pre-train\&Fine-tune      &$\textbf{74.3}$   &$\textbf{60.4}$  &$\times 1.16$\\ 
    \rowcolor{mygray}
     + STS Conv    &$\textbf{74.7}$ &$\textbf{61.4}$  &$\times 1.16$ \\ 
    \bottomrule
    \end{tabular}
    }
    \vspace{-0.5em}
    \label{tab:effectiveness_3x1x1}
  \end{minipage}
  \captionof{table}{\textbf{Effectiveness}. Ablation study on Image pre-training \& fine-tuning and Spatiotemporal separable 3D convolution (STS). We use 8 frames as input and all the models are pre-trained on ImageNet-1K.}
  \label{tab:effectiveness}
  \vspace{-2em}
\end{minipage}
\end{figure}

\paragraph{Effectiveness.} We first validate the effectiveness of the proposed method on K400 and SS-V2 in Table \ref{tab:effectiveness}. 
First, we control the total training budget to be close to from-scratch ($\times 1.16$). 
On the K400,
ResNet50-3$\times$1$\times$1 trained with our pipeline outperform the from-scratch by $\textbf{1.8\%}$.
For ResNet50-3$\times$3$\times$3  it improves the baseline by $\textbf{3.5\%}$.
Under the tight training computation,
image pre-training significantly improves the performance. 
It shows the importance of appearance modeling in video classification.
In addition,
replacing all the 3D convolution layer with STS convolution further improve the results.
On the SS-V2,
the proposed training pipeline can improve the accuracy by $\textbf{3.2\%}$ from \textbf{58.4\%} to \textbf{61.6\%} on ResNet50-3$\times$1$\times$1.
The same-scale improvement of $\textbf{2.9\%}$ can be found on ResNet50-3$\times$3$\times$3.
Moreover, 
on both backbones,
STS convolution notably improve the pre-trained models by an average of $\textbf{1.1\%}$
All the results demonstrate that under the same training budget, image pre-training can remarkably boost performance, and the STS convolution can also benefit from the appearance modeling prior to further improve performance.

\begin{table}[h!]
    \vspace{-2.0em}
    \centering
    \resizebox{0.95\linewidth}{!}{
    \begin{tabular}{c|c|c|c|c|c}
    \toprule
     ResNet50-\textbf{STS 3x3x3}&Input Frames & Pre-Train  &Fine-tune  & Training Budget &K400 Top1 \\
    \midrule
    \midrule
    From Scratch & 8 &-  &192M       &192M   &71.0 \\ 
    \rowcolor{mygray}
    ImageNet Pre. & 8 &128M   &96M       &$\times 1.16$   &\textbf{74.7} \\ 
    \midrule
    From Scratch & 16 &-  &384M        &384M   &73.5 \\ 
    \rowcolor{mygray}
    ImageNet Pre.   & 16   &192M   &192M     &$\times 1$  &\textbf{76.1} \\  
    \midrule
    From Scratch & 32 &-  &768M        &768M   &74.4 \\ 
    \rowcolor{mygray}
    ImageNet Pre.    & 32  &384M  &384M     &$\times 1$  &\textbf{76.7} \\  
   \bottomrule
    \end{tabular}
    }
    \vspace{0.5em}
    \captionof{table}{\textbf{More Frames \vs More Image Pre-training.} Following the fixed budget setting described in Section \ref{sec:budget}, we extend the pre-training epochs to match the total computation budget. }
    \label{tab:scaling}
    \vspace{-3em}
\end{table}

\paragraph{More input frames \vs more image pre-training epochs.} In this experiment, we aim to compare the capacity of the pre-training schedule when scaling up to different input frames. 
We use the ResNet50-3$\times$3$\times$3 equipped with STS convolution as the backbone. 
To make sure our total computation is consistent, 
we extend the image pre-training epochs from 100 epochs to 150 or 300 for 16 frames and 32 frames. 
The fine-tuning epochs remain the same for all models (50 epochs).
Surprisingly,
this simple scaling works well for more input frames.
As shown in Table \ref{tab:scaling}, employing image pre-training improve performance by $\textbf{3.7\%}$, $\textbf{2.6\%}$ and $\textbf{2.4\%}$ for 8, 16, and 32 frames, respectively. 
The stable performance gains demonstrate that our method can scale well over different input frames.
Furthermore,
these results may draw an important clue that why previous default pre-training pipeline does not work well: we should extend the pre-training schedule when our video models take more frames as input.

\begin{table}[h!]
    \vspace{-0.5em}
    \centering
    \resizebox{0.95\linewidth}{!}{
    \begin{tabular}{c|c|c|c|c|c}
    \toprule
     ResNet50-STS 3x3x3 &Input Frames & Pre-Train &Fine-tune  & Training Budget &K400 Top1 \\
    \midrule
    \midrule
    From Scratch & 8  &-  &192M       &192M   &71.0 \\
    \midrule
    \rowcolor{mygray}
    ImageNet Pre. & 8  &128M   &96M       &$\times 1.16$   &\textbf{74.7} \\ 
    \rowcolor{mygray}
    K400/Image Pre. & 8  &128M   &96M       &$\times 1.16$   & 73.4 \\
    \bottomrule
    \end{tabular}
     }
    \label{tab:csn frames and dataset}
    \vspace{-0.5em}
\end{table}

\begin{table}[h!]
    \centering
    \resizebox{0.95\linewidth}{!}{
    \begin{tabular}{c|c|c|c|c|c}
    \toprule
     X3D-S &Input Frames & Pre-Train &Fine-tune  & Training Budget &K400 Top1 \\
    \midrule
    \midrule
    From Scratch &  13 &-  &798M       &798M   &73.2 \\ 
    \midrule
    \rowcolor{mygray}
    ImageNet Pre. & 13 &384M   &312M       &$\times 0.87$   &\textbf{73.5} \\ 
    \rowcolor{mygray}
    K400/Image Pre. & 13 &384M   &312M       &$\times 0.87$   &72.8 \\
    \bottomrule
    \end{tabular}
    }
    \vspace{0.5em}
    \captionof{table}{\textbf{Choice of Pre-Training Dataset: ImageNet \vs K400}}
    \vspace{-3.em}
    \label{tab: x3d frames and dataset}
\end{table}

\paragraph{Source of image pre-training dataset.} We next discuss the influence of the dataset used for pre-training. 
We conduct experiments on two different training budgets (100 epochs \vs 256 epochs train from scratch). First, in the fixed budget setting, we use ResNet50-STS 3$\times$3$\times$3 as the backbone. Here we pre-train each frame of all videos in K400, and we match the total training computation with ImageNet by controlling the number of epochs, as shown in Table \ref{tab: x3d frames and dataset}. 
The performance of the ImageNet pre-train model is better than K400-Image pre-train under the same pre-training setting ($\textbf{74.7\%}$ \vs $\textbf{73.4\%}$). The k400-image pre-train model also has notable improvement over training from scratch from $\textbf{71.0\%}$ to $\textbf{73.4\%}$, proving the effectiveness of image pre-training. 
In the train for state-of-the-art setting,  the results obtained by pre-training on K400-image are slightly lower than from-scratch by $0.5\%$. In both settings, the ImageNet pre-training is consistently better than the K400-image. We assume that the label of the K400-image is still an action label, which can lead to confusion. Hence, we choose ImageNet as our source of image pre-training dataset by default.

\subsection{Ablation on Spatiotemporal Fine-Tuning}
\label{sec: Ablation_spatiotemporal_Fine-Tuning}
\begin{figure}[h!]
\vspace{-2em}
\begin{minipage}{\linewidth}
   \begin{minipage}{0.33\linewidth}
    \centering
    \resizebox{\linewidth}{!}{
    \small
    \begin{tabular}{c|c|c}
    \toprule
    ResNet50-\textbf{3x1x1}     &K400  & SS-V2   \\
    \midrule
    \midrule
    Inflation        &74.5    &61.8  \\ 
    \rowcolor{mygray}
     + STS Conv    &$\textbf{74.8}$ &$\textbf{62.6}$   \\ 
     \midrule
    Zero-Init        &74.9    &61.6  \\ 
    \rowcolor{mygray}
     + STS Conv    &$\textbf{75.0}$ &$\textbf{62.7}$   \\ 
    \bottomrule
    \end{tabular}
    }
    \label{tab:zero vs inflate 3x1x1}
  \end{minipage}
   \hfill
   \begin{minipage}[t]{0.32\linewidth}
     \centering
     \resizebox{\linewidth}{!}{
     \small
     \begin{tabular}{c|c|c}
     \toprule
     ResNet50-\textbf{3x3x3}     &K400  & SS-V2   \\
     \midrule
     \midrule
     Inflation         &74.0   &60.0  \\
     \rowcolor{mygray}
      + STS Conv    &$\textbf{74.6}$ &$\textbf{61.7}$   \\ 
     \midrule
     Zero-Init       &74.3   &60.4  \\ 
     \rowcolor{mygray}
      + STS Conv    &$\textbf{74.7}$ &$\textbf{61.4}$   \\ 
     \bottomrule
     \end{tabular}
     }
    \label{tab:zero vs inflate 3x3x3}
   \end{minipage}
   \hfill
   \begin{minipage}[t]{0.32\linewidth}
    \centering
    \small
    \resizebox{\linewidth}{!}{
    \begin{tabular}{c|c|c}
    \toprule
    ResNet50-\textbf{3x3x3}     &K400  & SS-V2   \\
    \midrule
    \midrule
      ($1/3, 1/3, 1/3$)   &74.0   & 60.0 \\ 
     ($1/3, 1/3, 1/9$)    &74.3   &60.0 \\ 
      ($1/9, 1/3, 1/9$)    &74.3 &60.2  \\ 
      \rowcolor{mygray}
      ($0, 1, 0$)   &$\textbf{74.3}$ &$\textbf{60.4}$   \\ 
    \bottomrule
    \end{tabular}
    }
    \label{tab:inflation rate 3x3x3}
  \end{minipage}
\vspace{-.5em}
\captionof{table}{\textbf{Different Initialization Strategies.}}
\label{fig:diff_init}
\vspace{-2em}
\end{minipage}
\end{figure}

\paragraph{Initialization strategy.} Next, we explore how to initialize the model in the fine-tuning phase. Inflation averages the pre-trained 2D model parameters along the temporal dimension \cite{k4002017}. We can initialize the center kernel of 3D convolution with pre-trained weights named as zero-init. Remaining parts are zeros. As shown in Table \ref{fig:diff_init}, in most cases, the zero-init approach exceeds inflation. 
ResNet50-3$\times$3$\times$3 zero-init exceeds the inflation by \textbf{0.3\%} on K400 and \textbf{0.2\%} on SS-V2.
ResNet50-3$\times$1$\times$1 zero-init outperforms the inflation by \textbf{0.4\%} on K400 but slightly lower on SS-V2.
These results suggest 3D CNNs prefer 2D image initialization. 
More importantly, our proposed STS convolution achieves a considerable improvement regardless of the initialization. 
For example, ResNet50-STS 3$\times$3$\times$3 improves the inflation baseline by \textbf{0.6\%} on K400 and \textbf{1.7\%} on SS-V2.
Compared with zero-init, it boosts the performance by \textbf{0.4\%} on K400 and \textbf{1.0\%} on SS-V2.
Same scale improvement can be found on ResNet50-STS 3$\times$1$\times$1.
It indicates that STS convolution can be optimized very well using different initialization. 
The performance improvement of STS convolution comes more from the specific design.
In the right table, we also evaluate different inflation rates, closer (\eg, (1/9, 1/3, 1/9)) to zero-init rate works better. 
It also indicates that the initialization strategy is less important than training pipeline setting.

\begin{figure}[h!]
\vspace{-1.0em}
\begin{minipage}{\linewidth}
   \begin{minipage}[t]{0.48\linewidth}
    \centering
    \resizebox{\linewidth}{!}{
    \begin{tabular}{c|c|c|c}
    \toprule
    ResNet50-3x3x3 &ImageNet-1K  &K400  &SS-V2 \\
    \midrule
    \midrule
    Baseline  &76.6  &74.3       &60.4  \\ 
    \midrule
    \rowcolor{mygray}
    w/o splitting  &76.6  &74.0       &60.8   \\ 
    \rowcolor{mygray}
    w/ splitting  &\textbf{77.1}    &\textbf{74.7}     &\textbf{61.4}  \\ 
    \bottomrule
    \end{tabular}
    }
    \vspace{0.5em}
    \label{tab:STS Conv}
  \end{minipage}
  \hfill
   \begin{minipage}[t]{0.48\linewidth}
     \centering
     \resizebox{\linewidth}{!}{
     \begin{tabular}{c|c|c|c}
     \toprule
     ResNet50-3x3x3 &ImageNet-1K  &K400  &SS-V2 \\
    \midrule
    \midrule
    1:1     &77.1    &\textbf{74.7}     &61.4 \\  
    1:2   &76.8  &74.6   &\textbf{61.8}  \\ 
    2:1   &\textbf{77.2}  &74.7   &61.4 \\ 
    \bottomrule
        \end{tabular}
        }
        \label{tab:STS Conv}
   \end{minipage}
   \vspace{-1em}
\captionof{table}{\textbf{Details of Design in STS Convolution.}}
\label{tab:ablation STS Conv}
\vspace{-2em}
\end{minipage}
\end{figure}

\paragraph{Ablation on STS convolution.} 
Next, we perform ablation on the STS convolution design. The experimental results are shown in Table \ref{tab:ablation STS Conv}. First of all, our baseline model is ResNet50-3$\times$3$\times$3. From the table on the left, we can see that the key factor in STS is whether to use splitting or not. 
Splitting operation utilizes all the weights in the original 3D convolution, increasing the spatial perception field. 
It can boost the video classification performance as same as image classification.
With splitting operation, it improves the without reshaping counterpart on the ImageNet (\textbf{+ 0.5\%}), K400 (\textbf{+0.7\%}) and SS-V2 (\textbf{+1.0\%}). On the right table, we examine how many channels should be divided. For example, ratio of $1:2$ means $1/3$ channels are used for appearance modeling. The results are very close on the K400. Thus we use $1:1$ as the default ratio.

\section{Conclusion}
We argue that image pre-training has significant potential for 3D video recognition, and should be more commonly adopted since 1) image pre-training models which are widely and freely available, contain enriched appearance information 
for facilitating temporal reasoning
2) image pre-training enables a large speedup for existing 3D CNN training.
Specifically, in this paper, we reveal that the key to exploiting image pre-training lies in properly decomposing the spatiotemporal features into spatial and temporal parts.
Based on this principle, we design a new image pre-training and spatiotemporal fine-tuning strategy, which achieves notable performance improvements for a wide range of 3D CNNs on multiple video recognition tasks with significant speedup. 

\section*{Acknowledgement}
We thank Chao-yuan Wu and Zeyu Wang for their insightful comments and
suggestions. This work is partially supported by ONR N00014-21-1-2812.

\bibliographystyle{splncs04}
\bibliography{egbib}
%
%

\end{document}


\appendix
\section*{Appendix}

This appendix provides further details as referenced in the main paper: 
Section \ref{STS formula} contains detailed description of proposed STS conv.  Section \ref{Additional} contains further results ablations on Kinetics-400.

\section{Formula of STS Conv} 
\label{STS formula}
We give the formal definition of STS convolution as following. Given the input $\mathbf{x}\in\mathbb{R}^{C\times T \times  H\times W}$ and a 3D Conv with weights $\mathbf{\theta}\in\mathbb{R}^{C_{in}\times C_{out} \times K_t \times  K_h \times K_w}$ (for simplicity, $C_{in} = C_{out} = C, K_t=3$),
We first decomposes the $\theta$ along the channel dimension  into two groups: $(\mathbf{\alpha} ,\mathbf{\beta} )\in\mathbb{R}^{C\times C_{1/2} \times 3 \times  K_h \times K_w}$.
$\alpha$  is for the static appearance modeling so we can split it along temporal dimension into $(\alpha_{0}, \alpha_{1}, \alpha_{2}) \in\mathbb{R}^{C\times C_{1/2} \times  K_h \times K_w}$.
$\beta$ is for dynamic motion modeling.
To preserve $\alpha_{1}$'s appearance modeling ability,
we initialize the $\alpha_{0}$ and $\alpha_{2}$ with zeros.
Then we aim at leveraging the \emph{untouched} $\alpha_{0}$ and $\alpha_{2}$ to enlarge spatial receptive field.
Specifically,
we reshape each frame ${x_{t}}$ into ${x_{t}^{row}}$ with size of $(C, W\times H)$ and ${x_{t}^{col.}}$  with size of $(C, H\times W)$. 
Similarly,
$\alpha_{0}$ and $\alpha_{2}$ should be reshaped.
Finally,
we gather the feature
\begin{align}
    \mathbf{y}_{t}
=\text{\textit{\textbf{\textrm{concat}}}}(
 &\underbrace{\textcolor{blue}{
{Conv1D}({x_{t}^{row}}; \alpha_{0})
+{Conv2D}({x_{t}}; \alpha_{1})
+{Conv1D}({x_{t}^{col.}}; \alpha_{2})}} 
_{\textcolor{blue}{Static}} \\
&,\underbrace{ \textcolor{red}{
Conv3D
(\mathbf{x};\beta)}}_{\textcolor{red}{Dynamic}}).   
\end{align}

\section{Additional Ablation Study}
\label{Additional}
\subsection{Case Study of Slowfast}
We believe that a proper initialization method and training schedule are the two keys to boosting 3D CNNs' performance.
First, 
we pre-train the two branches together while SlowFast only initializes the slow branch
due to its structural changes.
As shown in Table~\ref{tab:slowfast multi-grid},
pre-training both branches with STS improves SlowFast pipeline by \textbf{0.3\% }on the same amount of budget.
Second, further increasing the pre-training budget to 300 epochs readily outperforms the from-scratch result by \textbf{1.3\%} with only $\mathbf{\times 0.8}$  computation. 

\begin{table}[h!]
    \vspace{-1em}
    \centering
    \resizebox{\linewidth}{!}{
    \begin{tabular}{c|c|c|c|c|c}
    \toprule
    Model &Pre-train Branch &Pre-train  &Fine-tune  &Total Budgets &K400   \\
    \midrule
    \midrule
    SlowFast (from scratch) & -   &-  &256  &$\times1$  &75.6       \\ 
   SlowFast (previous pipeline) & slow   &90  &100 &$\times0.5$   &75.4          \\ 
      \midrule
    \rowcolor{mygray}
   STS-SlowFast (our pipeline) &slow+fast  &90  &100   &$\mathbf{\times0.5}$ &\textbf{75.7}      \\
   \rowcolor{mygray}
   STS-SlowFast (our pipeline) &slow+fast  &300  &100  &$\mathbf{\times0.8}$ &\textbf{76.9}       \\

    \bottomrule
    \end{tabular}
    }
    \vspace{1em}
\captionof{table}{Investigation of pre-training in SlowFast 4$\times$16 . 
}
\label{tab:slowfast multi-grid}
 \vspace{-3em}
\end{table}

\subsection{Dilated Conv v.s. STS Conv }
During fine-tuning, reshaping the \emph{untouched} kernels in spatial space can enlarge the receptive field to boost performance.
Two reshaped 1D Convs can obtain larger receptive filed than two same-directional dilated Convs. 
We ablate dilated Conv and have two observations. 1) Reshaping 1D Conv achieves better results than dilated Conv on SSV2 (61.4\% \vs 61.1\%) and K400 (74.7\% \vs 74.5\%). 2) Dilated Conv is better than the baseline (61.1\% \vs 60.4\% on SSV2, 74.5\% \vs 74.3\% on K400), suggesting the effectiveness of enlarging receptive field in the static channel. 
\begin{table}[h!]
 \vspace{-1.5em}
    \centering
    \resizebox{\linewidth}{!}{
    \begin{tabular}{c|c|c|c|c}
    \toprule
    ResNet50-3x3x3 &dilated rate &Effective Receptive field  &K400  &SS-V2 \\
    \midrule
    \midrule
    Baseline  &- &  $3\times3$  &74.3       &60.4  \\ 
    \midrule
    \rowcolor{mygray}
    w/ dilated conv  & 2   &$3\times3+5\times5$   &74.5    &{61.2}  \\ 
    \rowcolor{mygray}
    w/ dilated conv   &3   &$3\times3+7\times7$   &74.4     &61.1  \\ 
    \rowcolor{mygray}
    w/ two orthogonal 1D convs &-   &$1\times9+3\times3+9\times1$   &\textbf{74.7}     &\textbf{61.4}  \\ 
    \bottomrule
    \end{tabular}
    }
    \vspace{0.5em}
\captionof{table}{Dilated Conv v.s. STS Conv.}
\label{tab:dilated vs STS Conv}
 \vspace{-3em}
\end{table}